\documentclass[runningheads]{llncs}

\usepackage[T1]{fontenc}
\usepackage{graphicx}
\usepackage{hyperref}
\usepackage{color}

\usepackage{expex}

\begin{document}

\title{Word Overuse and Alignment in Large Language Models:\ The Influence of Learning from Human Feedback}

\titlerunning{Word Overuse, LLMs, and LHF}
%

\author{Tom S.\ Juzek\inst{1}\orcidID{0000-0002-3204-3879} \and Zina B.\ Ward\inst{1}\orcidID{0000-0003-0160-6656}\thanks{Conceptualization:\ TSJ, ZBW (eq.). Code, Methodology:\ TSJ. Write-up:\ TSJ, ZBW (eq.). GitHub repository:\ \href{https://github.com/tjuzek/lhf}{github.com/tjuzek/lhf}. Computational setup:\ 2024 Thelio Custom machine, GeForce RTX 3090.}}


\authorrunning{T.\ S.\ Juzek \& Z.\ B.\ Ward}

\institute{Florida State University, Tallahassee FL 32306, USA\\
}

\maketitle

\begin{abstract}
Large Language Models (LLMs) are known to overuse certain terms like ``delve'' and ``intricate.'' The exact reasons for these lexical choices, however, have been unclear. Using Meta's Llama model, this study investigates the contribution of Learning from Human Feedback (LHF), under which we subsume Reinforcement Learning from Human Feedback and Direct Preference Optimization. We present a straightforward procedure for detecting the lexical preferences of LLMs that are potentially LHF-induced. Next, we more conclusively link LHF to lexical overuse by experimentally emulating the LHF procedure and demonstrating that participants systematically prefer text variants that include certain words. This lexical overuse can be seen as a sort of misalignment, though our study highlights the potential divergence between the lexical expectations of different populations – namely LHF workers versus LLM users. Our work contributes to the growing body of research on explainable artificial intelligence and emphasizes the importance of both data and procedural transparency in alignment research.

\end{abstract}


\section{Introduction}
\label{sec:introduction}

Following the arrival of Large Language Models (LLMs), observers were quick to note their tendency to overproduce certain lexical entries \cite{Koppenburg2024,Nguyen2024,shapira2024delving,gray2024chatgpt,kobak2024delving,liang2024mapping,liu2024towards,matsui2024delving,juzek2025does}. Much of the discourse centered on Scientific and academic English, focusing on words such as ``delve'', ``intricate'', and ``realm.'' For this reason, we also concentrate on Scientific English here. While changes in Scientific English over decades and centuries are well-documented \cite{degaetano2018using,degaetano2018information,bizzoni2020linguistic,menzel2022medical}, the language shifts following the introduction of LLMs have been unprecedented, with certain words (like ``delve'') seeing a sudden and dramatic increase in usage. 

\begin{figure}[ht]
    \centering
    \includegraphics[width=0.7\columnwidth]{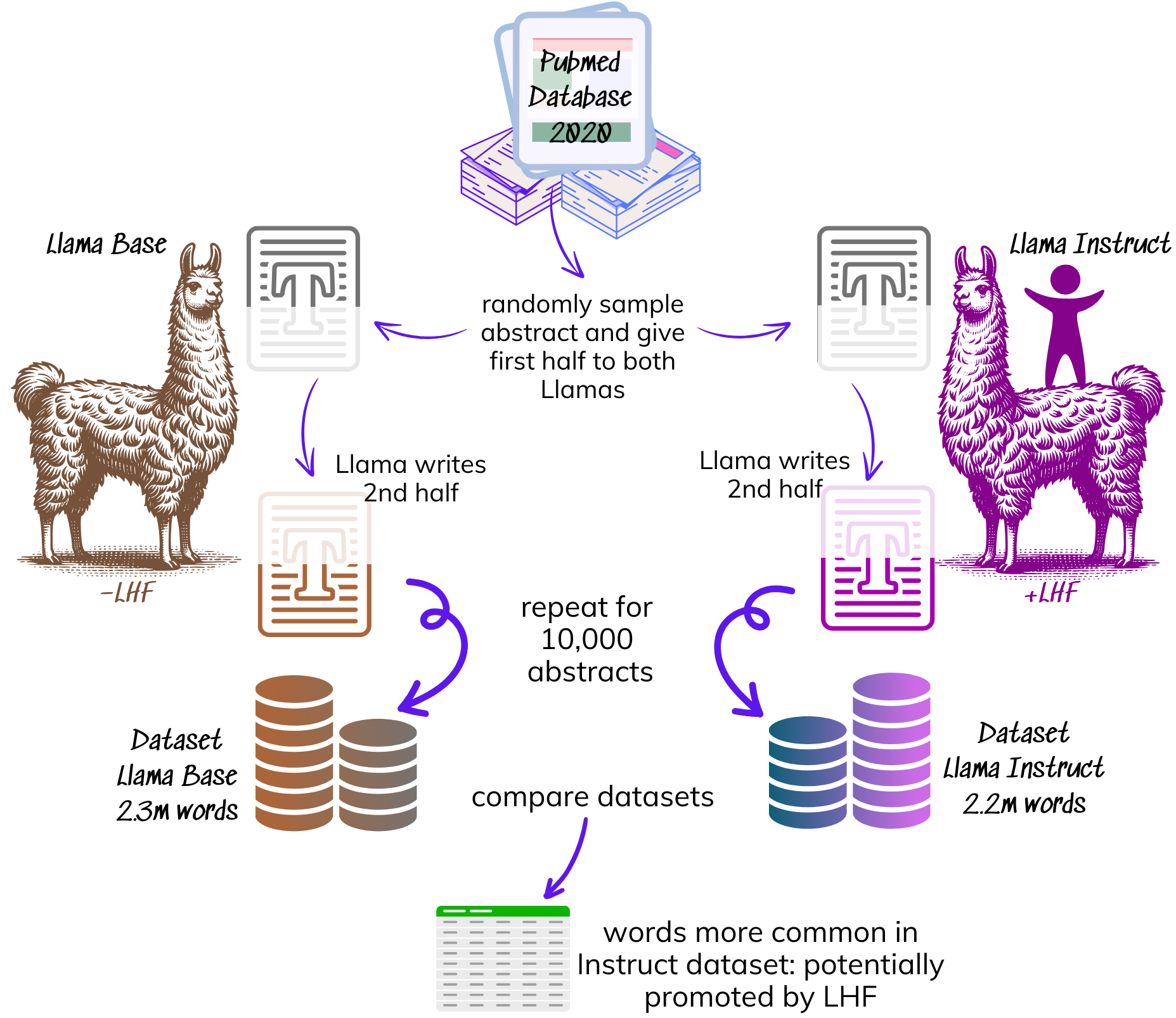}
    \caption{An illustration of the procedure used to identify lexical preferences that are potentially induced by Learning from Human Feedback (LHF); created with Canva. \vspace{-0.5cm}}
    \label{fig:identificationprocedure}
\end{figure}

Thus, it has been established \textit{that} certain lexical biases exist in LLMs, with evidence demonstrating their influence on written language. However, the question of \textit{why} this lexical overrepresentation arises remains open. While some have pointed to Learning from Human Feedback (LHF) as a significant contributor to these lexical choices \cite{Hern2024,sheikh2024delve}, conclusive evidence is still missing. 

Learning from Human Feedback is a procedure applied after initial model training during which human evaluators indicate preferences through A/B testing or ranking. It was first introduced in the form of Reinforcement Learning from Human Feedback (RLHF; \cite{christiano2017deep,ziegler2019fine}), though a more recent and increasingly popular form of LHF is Direct Preference Optimization (DPO), which aligns models by directly optimizing for human preferences without relying on reinforcement learning \cite{rafailov2024direct}. LHF was introduced to align models more closely with human preferences. Alignment, which reflects ``how closely the model's opinions or stances mirror those of different social groups'' \cite{he2024whose}, is a major challenge in AI \cite{bender2021dangers,santurkar2023whose,durmus2023towards}. A model is \textit{misaligned} for a target group when its output does not align with the group's opinions, values, and/or expectations. LHF is recognized as a key factor contributing to the success of models like ChatGPT \cite{ouyang2022training}. However, researching the effects of LHF is difficult due to lack of transparency surrounding the procedures and datasets used in model development \cite{bommasani2021opportunities}. 

The present study addresses the potential link between LHF and the lexical choices of LLMs through a two-step process. First, we introduce a method for identifying lexical preferences in LLMs that are potentially induced by LHF. This procedure can aid efforts to mitigate the most extreme cases of lexical overrepresentation (Section~\ref{sec:procedure}). Second, we conduct an experiment that emulates the LHF procedure in order to test whether humans indeed prefer texts containing the words identified by our initial procedure. This represents an empirical test of the hypothesis that LHF plays a role in shaping LLMs' lexical choices (Section~\ref{sec:experiment}), based on one of Meta's popular Llama models. While our findings provide evidence for an LHF effect, other contributing factors remain to be systematically investigated. Finally, we discuss the implications of our study (Section~\ref{sec:discussion}) and its limitations (Section~\ref{sec:limitations}). 

\subsection*{Related Work}
\label{sec:related}

Many studies explore the linguistic behavior of LLMs and their effects on (written) human language \cite{Koppenburg2024,Nguyen2024,shapira2024delving,gray2024chatgpt,kobak2024delving,liang2024mapping,liu2024towards,matsui2024delving}, with a few investigating spoken language \cite{geng2024impact,yakura2024empirical}. Most of this work is situated at the word level, though there is also research  on syntactic behavior \cite{zamaraeva2025comparing}. Procedures for identifying LHF-induced overuse have been proposed \cite{kobak2024delving,juzek2025does}, but these involve a manual component \cite{kobak2024delving,juzek2025does} and/or have a different focus \cite{kobak2024delving}. Similar concerns have been raised for other behaviors exhibited by LLMs (see the xAI literature; \cite{sculley2015hidden,zhao2024explainability,cambria2024xai}). Overlap between human linguistic preferences and model behavior has been shown, though with a small sample  \cite{juzek2025does}. For non-lexical form (such as boldface or emoji use), it has been found that even subtle differences in preferences during human preference training can result in substantial differences in  model behavior \cite{zhang2024lists}.

\section{Procedure to Identify Potentially LHF-Induced Lexical Preferences}
\label{sec:procedure}

As a first step, we develop a low-cost procedure to identify lexical preferences in LLMs that may originate from LHF training. Our approach involves generating language outputs from both a pre-LHF model and a post-LHF model and then comparing word usage in the outputs. Here, we use Llama 3.2-3B Base and Llama 3.2-3B Instruct \cite{dubey2024llama} (via the Hugging Face Transformers library \cite{wolf2020transformers}). The Llama family is, to our knowledge, the closest available approximation between models trained with and without LHF, which for Llama 3 involves Direct Preference Optimization. At the time of our research, Llama 3.2 was the most recent version of the Llama model family. While larger variants (11B and 90B) were available, they primarily added multimodal capabilities; improvements in textual reasoning abilities were minor \cite{huggingface2024leaderboard}. At the time of research, a broader model comparison would have been difficult:\ of all the major LLM developers, only Meta had released both base and instruction-tuned models. Since then, models like OLMo \cite{allenai2024olmo2} and Falcon \cite{tii2024falcon3} have gained popularity and would now be strong options. 

There are other differences between Llama Base and Llama Instruct \cite{dubey2024llama}, most notably instruction tuning, optimization for tooling, and safety mitigation. However, none of these, including instruction tuning \cite{juzek2025supplement}, are known to contribute to lexical overrepresentation. LHF remains the most plausible contributor to shifts in language output. This makes the Llama models well-suited for our purposes. All technical implementations described in this paper were carried out in Python 3 (\cite{Python3}; v3.12.3).

Although our study focuses on Scientific English, the procedure we present is transferable to other domains. Here the procedure is applied to abstracts from PubMed from 2020 \cite{PubMed}, as this predates the mainstream availability of LLMs. We randomly sampled 10~000 abstracts and filtered out those with fewer than 40 words, which resulted in 9 853 abstracts. Each abstract was split in half by word count (rounding down), and each of the Llama models, Base and Instruct, were prompted to continue writing based on the initial half of the abstract (Prompt:\ `Continue the following academic article:\ $\backslash$``\{first\_half\} '). Models were, if needed, cut off after twice the input length. The generated continuations were cleaned in order to remove issues such as generation loops (e.g.,\ repetitive sentences) and meta-comments (e.g.,\ ``Certainly, here is ...''), using GPT-4o \cite{achiam2023gpt,openai_python_api} (Prompt:\ `The following text is meant to be a continuation of a scientific abstract. In some of the continuations, however, the AI finishes the abstract and continues with commentary. Please detect potential switches, and remove any commentary:\ $\backslash$n$\backslash$n``\{input\_text\}''$\backslash$n$\backslash$n Output only the cleaned abstract. If the entire text is commentary, output an empty string.'). 

This process resulted in two corpora of PubMed abstract continuations:\ one generated by Llama Base (totaling 2.3m words) and the other by Llama Instruct (2.2m words). Both corpora were tagged for part-of-speech using spaCy (\cite{montani2023spacy}; v3.8.3, en\_core\_web\_sm v3.8.0, tagging of all data took about 140hrs), enabling the disambiguation of identical surface forms across word categories (e.g.,\ ``to\_PART run\_VERB'' vs.\ ``a\_DET run\_NOUN'') and the grouping of conceptually related forms under a common lemma (``delve'' and ``delves''). Relative frequency usage was compared between the two corpora (similar to what one sees in the Google Ngram Viewer \cite{google2024ngram}). Here and in Section~\ref{sec:experiment}, we focus on statistically significant differences between Base and Instruct lexical usage, determined through a chi-square test. The top five items showing an increase in usage in the Instruct model compared to the Base model are as follows:\ ``nuanced\_ADJ (+8342\%)'', ``nuance\_VERB (+6301\%)'', ``firstly\_ADV (+4794\%)'', ``reliance\_NOUN (+3193\%)'', ``generalizability\_NOUN (+3124\%)''; also see Table~\ref{tab:buzzwords} for further entries and our GitHub for the full list. 

\begin{table}[ht]
\centering
\begin{tabular}{|l|c|c|c|}
\hline
\textbf{Lemma\_POS} & \textbf{opm Ll-B} & \textbf{opm Ll-I} & \textbf{Incr.\ \%} \\ 
\hline
nuanced\_ADJ  &  0.6  &  51.4  &  8342.8  \\
nuance\_VERB  &  0.6  &  39  &  6301.7  \\
firstly\_ADV  &  2.4  &  119.2  &  4794  \\
reliance\_NOUN  &  1.2  &  40.1  &  3193.6  \\
generalizability\_N  &  2.4  &  78.5  &  3124  \\
underscore\_VERB  &  4.3  &  124.9  &  2829.1  \\
radar\_NOUN  &  0.6  &  16.4  &  2590.6  \\
staffing\_NOUN  &  0.6  &  13  &  2033.9  \\
socioemotional\_ADJ  &  0.6  &  13  &  2033.9  \\
multifacete\_VERB  &  0.6  &  11.9  &  1848.3  \\
flake\_NOUN  &  0.6  &  10.7  &  1662.8  \\
interoceptive\_ADJ  &  0.6  &  10.7  &  1662.8  \\
vocabulary\_ADJ  &  0.6  &  10.7  &  1662.8  \\
theanine\_NOUN  &  0.6  &  10.7  &  1662.8  \\
secondly\_ADV  &  6.1  &  103.4  &  1597.8  \\
finish\_NOUN  &  0.6  &  10.2  &  1570  \\
daa\_NOUN  &  0.6  &  10.2  &  1570  \\
necessitate\_VERB  &  0.6  &  9.6  &  1477.2  \\
behavioral\_NOUN  &  0.6  &  9.6  &  1477.2  \\
\hline
\end{tabular}
\caption{Lemmata and part-of-speech for the Top 20 words identified using the procedure described in Section~\ref{sec:procedure}. Compared are occurrences-per-million (opm) for Llama Base (Ll-B) vs.\ Llama Instruct (Ll-I).}
\label{tab:buzzwords}
\end{table}

Our procedure serves as a proof of concept:\ the identification of lexical items potentially favored by LHF can be automated. The procedure is validated in part by the observation that many of the identified words have been discussed in the literature on the distinctive lexical choices of LLMs \cite{shapira2024delving,gray2024chatgpt,kobak2024delving,liang2024mapping,liu2024towards,matsui2024delving,juzek2025does}. However, the procedure does not necessarily identify words that are overused by Llama Instruct relative to human-generated text; the operative comparison is with Llama Base. Nevertheless, there seems to be considerable overlap between the words overused by Instruct relative to Base, and the words overused by Instruct relative to a human baseline. We compared the Llama Instruct outputs to a human baseline, the actual second halves of the randomly sampled PubMed abstracts. Almost all (813 out of 814) of the words used significantly more by Llama Instruct than Llama Base (Table~\ref{tab:buzzwords}) were also used significantly more by Instruct than in the human baseline. Thus, when it comes to the lexical items that distinguish LLM-generated text from human-generated text, the procedure in its current form effectively identifies many of the most extreme cases. 

Assuming such divergences from human-generated text are undesirable and hence a form of bias (a point to which we will return in Section~\ref{sec:discussion}), the procedure is a method for uncovering lexical biases in LLMs. The degree of such bias observed in LLM outputs suggests that either no robust identification mechanisms were  applied during model development, or existing mechanisms have proven too weak, which motivates the need for a procedure like ours. Our insights could also inform the discourse on AI-generated text detection \cite{lavergne2008detecting,chakraborty2023possibilities,mitchell2023detectgpt,huang2025magret}, as such methods often rely on identifying atypical lexical items and distributions. 

The above results are consistent with the hypothesis that LHF is a primary source of the lexical bias discussed in the literature. However, more conclusive evidence is needed; and specifically, experimental validation is required to confirm that the lexical items whose usage by LLMs we pinpointed as potentially LHF-induced are indeed preferred by human evaluators, thereby strengthening the causal link between LHF and LLMs' lexical choices.

\section{Experimental Validation}
\label{sec:experiment}

At the core of the hypothesized link between LHF and LLMs' lexical choices is the idea that evaluators exhibit a subtle preference for certain lexical items, a preference that is in fact so slight that it has obscured this very link. However, when scaled up, these minor preferences for specific lexical items become entrenched and ultimately manifested in the output generations of LLMs. To test this hypothesis, we created experimental items consisting of pairs of text variants. In each pair, one variant exhibits fewer words previously identified as potentially favored by LHF, while the other exhibits more such words, with all other factors held as equal as possible, including length and content. This design aims to isolate the effect of the presence of the lexical items identified above on evaluator judgments. 

\subsection{Experimental Setup}

\noindent \textbf{Creation of Experimental Items.} The ideal test of the hypothesis would involve creating two random variants of a given abstract, repeating this for tens of thousands of pairs, collecting human evaluations for all these pairs, and then analyzing the ratings. The problem, however, is that detecting the hypothesized subtle effect experimentally under this approach would require an extraordinarily high number of ratings to achieve statistical significance. Thus, we opted for a procedure that increases the lexical differences between items, while at the same time maintaining comparable validity and being less resource-intensive. 

For 50 randomly selected PubMed abstracts from 2020, we prompted GPT-4o to write summary notes (``The following text is an abstract from a scientific paper:$\backslash$n$\backslash$n\{input\_text\}$\backslash$n$\backslash$nSummarize the abstract in keywords, separate keywords by commas.''; see example on our \href{https://github.com/tjuzek/lhf/tree/main/appendices}{GitHub}). Using these summary notes as input, we then had Llama Instruct generate 500 abstracts (variants) for each item (Prompt:\ `Based on the following keywords, write a 100-word abstract for a scientific journal article:\ ``\{line\_of\_keywords\}.'' Reply with the abstract only.'), resulting in a total of 25~000 variants (50 random abstracts * 500 variants). We used GPT-4o to clean the abstracts (Prompt:\ `The following text contains a scientific abstract, but sometimes further text:$\backslash$n$\backslash$n``
\{input\_text\}''$\backslash$n$\backslash$nPlease remove any irrelevant text, which can include titles, incomplete sentences, even a comment that an abstract is to follow ($\backslash$``Abstract:\ $\backslash$''). Output only the cleaned abstract.'). We controlled for length by filtering out candidates that were below 90 or above 110 words. It has been widely recognized that ``delve'' is an LLM-associated word \cite{shapira2024delving,kobak2024delving,liu2024towards,matsui2024delving,juzek2025does} and a corresponding backlash against it \cite{juzek2025does}. Thus, we removed any variants containing any of the 21 most overused `AI words' as discussed in \cite{juzek2025does}, including words like ``realm'' and ``groundbreaking''. After applying these filters, our final set contained 8710 variants. 

For these items (also part-of-speech tagged), we calculated a score to measure a word's potential to have been favored by LHF (``LHF-Score''). Using the lexical items identified in Section~\ref{sec:procedure} as potentially promoted by LHF, we assigned a score to each variant by summing occurrences of these items, weighted by their relative rate of increase. This weighting reflects the idea that a single usage of a term like ``revolutionize\_VERB'', which experienced an increase of +1160\%, is probably more indicative of the influence of LHF than using a term like ``of\_ADP'', which saw an increase of only 2\%. As such, the score focuses on relative changes:\ A 100\% shift from 1 to 2 occurrences of a given word should be treated the same as a shift from 1000 to 2000 occurrences in that same token span.

The LHF-Score for a sequence is the sum of LHF-Scores for each token (\textit{w}). The LHF-Score for a given token is its increase in percent between Llama Base (\textit{B}) and Llama Instruct (\textit{I}), divided by one thousand (for ease of interpretability); ``opm'' stands for occurrences per million and is just the frequency of a token divided by the total number of tokens (\textit{N}), multiplied by one million. \vspace{-0.4cm}

\begin{eqnarray}
\mbox{LHF-Score}(S) &=& \sum_{i=1}^{n} \mbox{LHF-Score}(w_i) \nonumber \\
&& \mbox{\textit{where}} \nonumber \\
\mbox{LHF-Score}(w) &=& \frac{1}{1000} \cdot \left( \frac{\mbox{opm}_I(w) - \mbox{opm}_B(w)}{\mbox{opm}_B(w)} \times 100 \right) \nonumber \\
&& \mbox{\textit{where}} \nonumber \\
\mbox{opm}(w) &=& \frac{\mbox{count}(w)}{N} \times 10^6
\end{eqnarray}

\noindent An LHF-Score was calculated for all 8710 variants generated for the 50 summarized abstracts. For each of the 50 abstracts, we calculated the difference between the variant with the lowest LHF-Score and the one with the highest LHF-Score. We then selected the Top 30 abstract pairs with the largest Deltas while ensuring that the pair of variants were length-matched (in two cases, a length match was difficult, and we took the runners-up). The following hypothetical example between Sequence~\ref{seq:buzzwords} and Sequence~\ref{seq:baseline} illustrates how the LHF-Scores were calculated. The LHF-Score Delta is 0.31 (the score is calculated on lemmata and part-of-speech, which are omitted below for simplicity). A real example can be found on our \href{https://github.com/tjuzek/lhf/tree/main/appendices}{GitHub}. 

\ex \label{seq:buzzwords}
\begingl
\gla This is an intricate example full of complex words (SUM)//
\glb 0.03  0  0  0.36  0.03  0  0  0.2  0 (=0.44)//
\endgl
\xe

\vspace{-0.5cm}

\ex \label{seq:baseline}
\begingl
\gla This is a baseline example free from these words (SUM)//
\glb 0.03  0  0  0  0.03  0  0  0.07  0 (=0.13)//
\endgl
\xe

\noindent For the 30 selected items, the average LHF-Score for the variants with many of the lexical items identified in Section~\ref{sec:procedure} is 7.2 (average length:\ 105 words), and the average LHF-Score for the variants with the fewest such items is 1.7 (average length:\ 104 words). The complete set of experimental item pairs is available on our GitHub repository. A small number of the words identified by the procedure above do not seem likely to have been promoted by LHF, such as ``radar'' (see Section~\ref{sec:limitations}). This introduces noise into the experiment. For instance, one variant of an abstract might include ``radar'', resulting in a higher LHF-Score, even though the in- or exclusion of such a word is unlikely to affect human preference between the two variants. Such cases weaken the statistical power of the experiment and increase the risk of a false negative outcome (the beta rate), thereby favoring the null hypothesis \cite{haslwanter2016introduction}. We anticipate this effect to be minor, however, given that the majority of lexical items previously identified do seem plausibly the sort that are potentially promoted by LHF. \\

\noindent \textbf{Participants}. We recruited 400 participants (231 female, 169 male; average age:\ 30.1 years, standard deviation:\ 9.8) through Prolific (\href{https://www.prolific.com/}{www.prolific.com}). Tech companies often recruit LHF workers from the Global South \cite{kwet2019digital,perrigo2023exclusive,gray2024chatgpt,rohde2024broadening}. To more closely emulate the process by which LLMs are trained, we recruited participants from countries in the Global South where English is an official or widely used language (see Appendix~\ref{app:appendix} for a full list of countries). 90\% of our participants were from Africa and 10\% were from Southeast Asia. Participants were compensated at a rate equivalent to an average of \$15 per hour. \\ 

\begin{figure}[ht]
    \centering    \includegraphics[width=0.8\columnwidth]{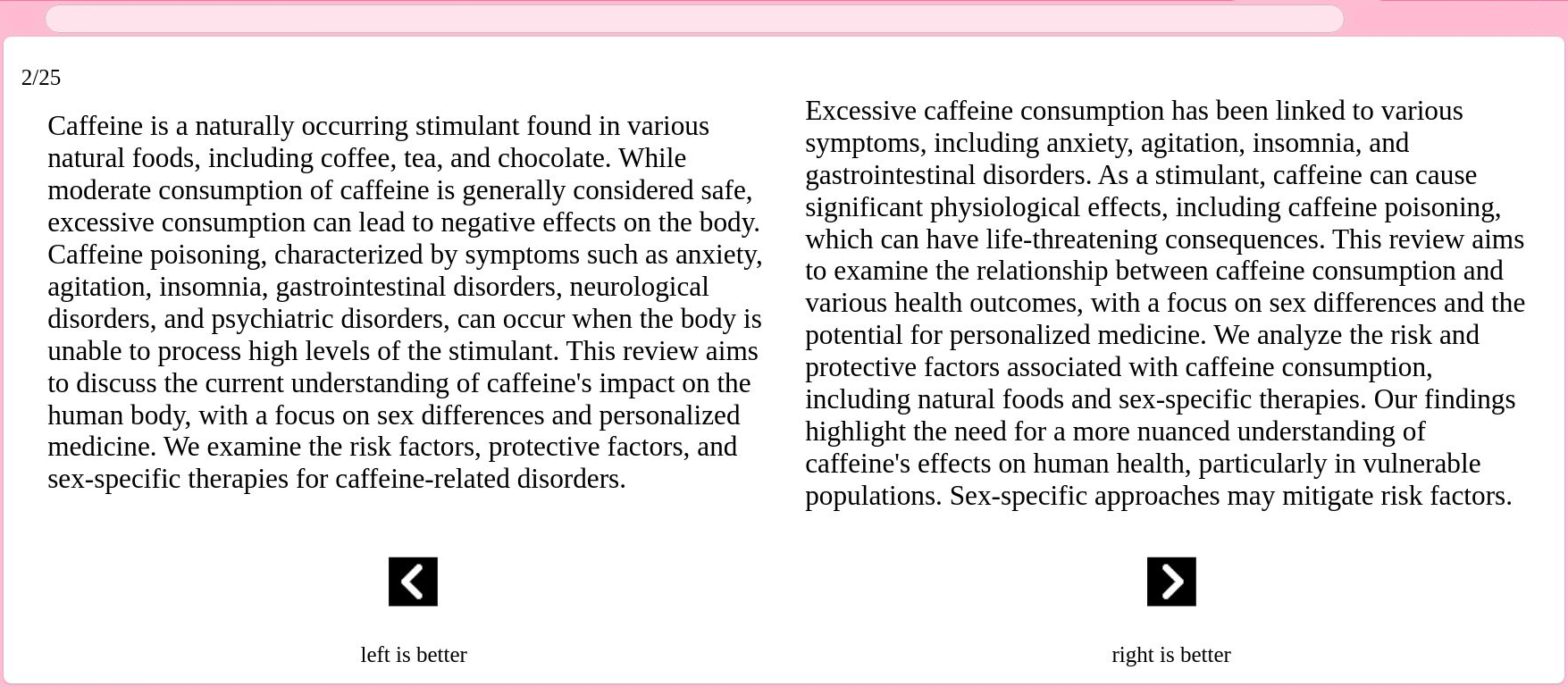}
    \caption{The rating interface for our experiment.}
    \label{fig:interface}
\end{figure}

\noindent \textbf{The Task.} The task began with IRB information (full instructions can be found on our \href{https://github.com/tjuzek/lhf/tree/main/appendices}{GitHub}), followed by an introduction to the task (``In the following, you will read a series of research summaries, with two alternatives next to each other. Please express which alternative you overall prefer. Some of the items are hard, do the best you can!'', with an example as per Figure~\ref{fig:interface}), including an example to familiarize participants with the process (for general best practices of experimental design, we followed \cite{cowart1997experimental} and \cite{berinsky2014separating}). Each participant rated 25 pairs of text variants, consisting of 20 critical item pairs (in random order), one calibration item at the beginning of the survey (where one variant was deliberately poor), two randomly interspersed ``gotcha'' items (which contained mid-sequence, \ ``This is not a real item, please click on the left button''; cf.\ \cite{berinsky2014separating,maniaci2014caring}), and two randomly interspersed items to assess language proficiency, similar to the calibration item. For each item, the left-right positioning of the abstracts was randomly flipped to avoid positional bias \cite{friedman1994biasing,chyung2018evidence}. We did not include fillers, as the differences between the variants were subtle, and we were not concerned that participants would guess the purpose of the study. 

\vspace{0.3cm}

\noindent \textbf{Exclusions.} To ensure high-quality data, which is crucial for statistical power \cite{mahowald2016snap}, we applied exclusions. Only participants who completed 10 or more of the 25 items were included in the analysis (11 participants excluded). Participants who failed to correctly answer both ``gotcha'' items were also excluded from the analysis (158 participants excluded). The literature reports that (225 ms + 25ms * character length of an item) is a good approximation of the minimum time physically required to read text \cite{haussler2017hot}. To account for skimming or decisions made on the basis of reading only part of each abstract, we used a less strict threshold, excluding only ratings completed in less than 40\% of this minimum time. Participants were warned if they responded more quickly than this. If a participant fell below this threshold on 5 or more items, all of their ratings were excluded from the analysis (18 additional participants excluded; many of the participants who failed the ``gotcha'' items would also have been excluded by this speed criterion). After exclusions, we retained 4039 ratings (out of a maximum of 8000 ratings:\ 400 participants * 20 ratings each), averaging about 135 ratings per item pair (minimum:\ 125 ratings). An exclusion rate of 46.8\% is in line with previous work \cite{downs2010your,zhu2010analysis,kazai2011worker,thomas2017validity,daniel2018quality}.

\subsection{Analyses}

The null hypothesis is that participants' choices between the high and low LHF-Score abstracts do not diverge from what one would expect when flipping a fair coin. The relevant alternative hypothesis is that participants show a preference for variants containing more of the words identified previously as potentially promoted by LHF – i.e., variants with a high LHF-score. For categorical, binary preference data like ours, where observations are tested against an expected baseline, a chi-square test is an excellent choice \cite{haslwanter2016introduction}. This is our main analysis. Additionally, we provide descriptives for the 30 item pairs, and we perform a mixed linear regression analysis to account for random effects. Our model includes the intercept as a fixed effect and participant and item as random effects. 

\begin{figure}[ht]
    \centering    \includegraphics[width=1\columnwidth]{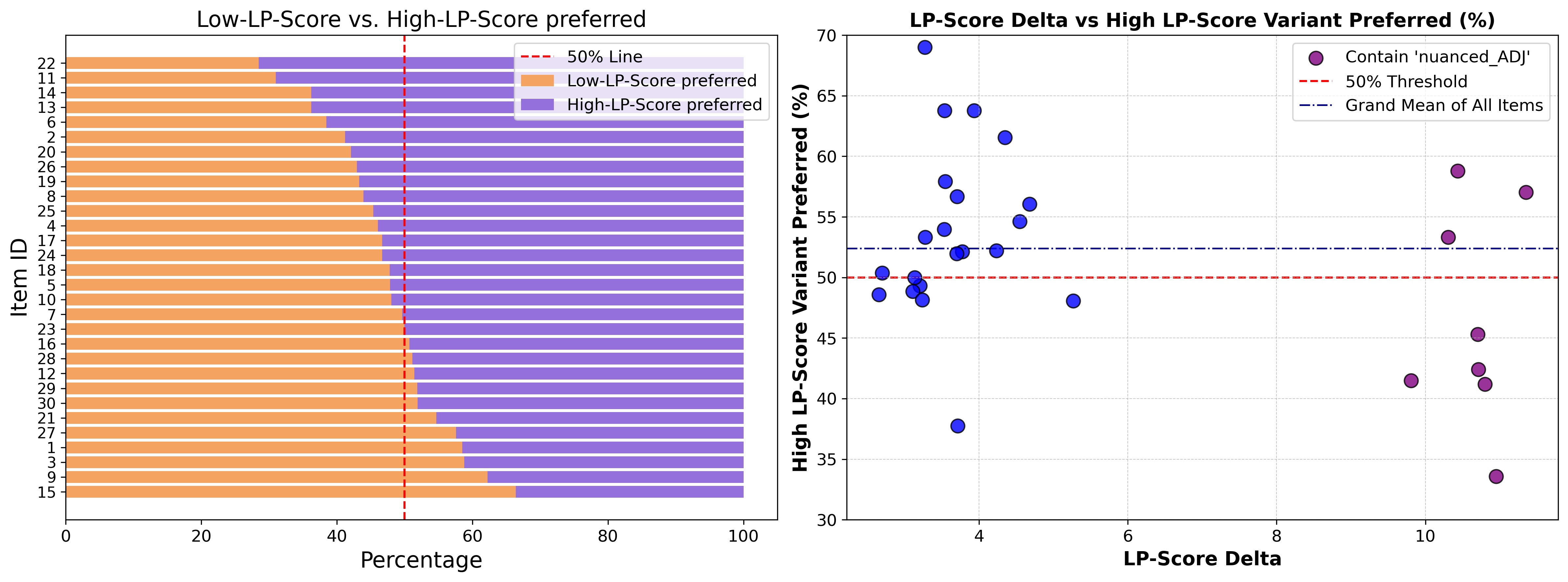}
    \caption{(a) Experimental results:\ Preferences between low LHF-Score variant vs.\ high LHF-Score variant, for the 30 items. (b) Participant preferences for pairs with different LHF-Score Deltas. Each dot represents the mean preference for one of 30 abstract pairs. High LHF-Score Delta pairs contained "nuanced\_ADJ."}
    \label{fig:itemresults}
\end{figure}

\subsection{Results}
\label{sec:results}

Overall, participants exhibited a highly significant preference for variants with a high LHF-Score over variants with a low LHF-Score (52.4\% to 47.6\%; \( \chi^2 = 9.4, p < 0.01 \)). This trend is consistent across items, as confirmed by the regression model and the low variance observed across items (also see Figure~\ref{fig:itemresults}). The mixed-effects model (REML, \( N = 4038 \), \(\mbox{log-likelihood} = -2903.53\)) revealed a significant intercept (\( \beta = 0.524, z = 33.20, p < 0.001 \)), with low variance across items (\( \sigma^2_{\mbox{item}} = 0.006 \)) and low to moderate variance across users (\( \sigma^2_{\mbox{user}} = 0.104 \)). Based on these findings, we reject the null hypothesis and accept the alternative hypothesis:\ participants systematically and significantly prefer variants containing more of the items identified in Section~\ref{sec:procedure} as words whose use by LLMs was likely promoted by LHF. 

Although we did not initially intend to analyze abstracts containing any particular word, we noticed that sentence pairs in which the high RP-Score abstract contains the adjective ``nuanced'' had a substantially higher LHF-Score Delta (Figure~\ref{fig:itemresults} (b)). Further, the average preference for the high LHF-Score variant is markedly lower for items containing ``nuanced'' (46.6\%) compared to sentence pairs without it (54.5\%). It could be that items containing ``nuanced'' stuck out to participants, leading them to disprefer those items, similar to what has been observed with text that includes ``delve'' \cite{juzek2025does}. Additional data is needed to substantiate this interpretation, however. 

\section{Discussion}
\label{sec:discussion}

It has been well established \textit{that} Large Language Models output certain words more frequently than a human baseline \cite{shapira2024delving,gray2024chatgpt,kobak2024delving,liang2024mapping,liu2024towards,matsui2024delving,juzek2025does}. Our research advances the discourse by addressing the \textit{why}, providing evidence that Learning from Human Feedback could be a primary source of this lexical overuse. We have identified lexical entries that models trained on LHF use considerably more than models without LHF training and then shown that texts containing many of these words are preferred to texts with fewer of them. 

Furthermore, there is reason to think that the words used more by Llama Instruct than by Llama Base are also the sorts of words overused by LLMs compared to humans. To probe this connection to human language use, we extracted the lexical entries discussed in the academic literature on lexical overrepresentation \cite{gray2024chatgpt,kobak2024delving,liang2024mapping,liu2024towards,matsui2024delving,juzek2025does}. This resulted in a list of 32 lexical entries (see Appendix~\ref{app:appendix}). We observe that 28 of these are also present in our Llama Base vs.\ Llama Instruct list. Thus, almost all of the words that researchers have identified as overrepresented in LLM-generated text compared to human-generated text appear more in the outputs of Llama Instruct than Llama Base. And as we have shown experimentally, these words are also favored by human evaluators, lending credibility to the hypothesis that the overuse of certain words by LLMs (relative to human usage) is at least partly the product of LHF. Our work therefore substantiates the previously speculative link between lexical overrepresentation and LHF.

It remains to be seen whether it is the demographics of the human evaluators or something about the feedback task they are engaged in that explains why they favor the sorts of words under discussion here. One notable observation is that LHF workers tend to be young, and almost all of the words overrepresented in LLM-generated text relative to human-generated text were already increasing in usage before the advent of LLMs \cite{matsui2024delving}. Taken together, these facts suggest that lexical overuse in LLMs might be a form of normal intergenerational language change \cite{labov2011principles}, albeit an accelerated one, wherein the preferences of younger generations are propagated in LLMs. This aligns with observations that young people tend to prefer AI-generated output over human-produced output \cite{young2024role}. 

LHF workers are also typically located in the Global South \cite{kwet2019digital,perrigo2023exclusive}, whereas criticism of the increased usage of words like ``delve'' has predominantly originated from the Global North. Most of the academic research on the topic, such as \cite{gray2024chatgpt,kobak2024delving,liang2024mapping,matsui2024delving,juzek2025does}, has been conducted at institutions based in the Global North. Some have speculated that the words overrepresented in LLM outputs might be more common in the dialects of English spoken by these LHF workers \cite{Hern2024,sheikh2024delve}, though follow-up work has not yet substantiated this conjecture \cite{juzek2025does}. 

It is also possible that it is the nature of the LHF task that is responsible instead. Perhaps human evaluators, skimming quickly through unfamiliar text, rely on the presence of certain words as a proxy for quality. It was shown that human evaluators tend to prioritize style over content \cite{wu-aji-2025-style}, which may explain why evaluators treat certain words as indicative of good outputs. In that case, the lexical preferences baked into LLMs through LHF might simply be task-driven. Discriminating between these explanations – that is, determining whether age, geographic location, dialect, or task features lead LHF workers to favor particular words – requires future research.

\section{Limitations}
\label{sec:limitations}

This work has several limitations. First, our analysis is restricted to Meta's Llama. Broader validation would require access to base and instruction-tuned model variants from other LLM developers (such as OLMo or Falcon). Our analysis also focuses on English. Expanding this work to other languages would be valuable. Furthermore, while our dataset contains approximately 2m tokens per model, future work could scale this up. A likely artifact of the corpus size is the occasional identification of lexical items that are not commonly cited as overused by LLMs. For instance, the Instruct model uses the item ``radar\_NOUN'' considerably more often than the Base model (+2590\%). A qualitative analysis of the dataset, however, helps to make sense of this result:\ several PubMed abstracts in our sample discuss ``radar\_NOUN'', and the Instruct model incorporates this into its continuations, whereas the Base model does not. Thus, scaling our procedure could improve the results. 

Potential language confounds in the experimental items might have impacted our results. While we controlled for abstract length, other distinctive linguistic features of LLM-generated text, such as specific syntactic structures or stylistic elements (e.g., ``It's not about [X], it's about [Y]'' \cite{jim2024aiwriting}), might correlate with the presence of the words that we have identified, unknowingly contributing to higher preference ratings. A qualitative inspection of the item pairs did not reveal any clear patterns of such confounding features, but the possibility cannot be entirely ruled out. Furthermore, although our experimental procedure aimed to emulate the task situation of LHF workers, it did so imperfectly, as we cannot perfectly simulate their working conditions for both ethical and practical reasons. Lastly, while our experimental results clearly bear on the existing discourse about lexical biases, the connection to human language use remains somewhat preliminary. Further strengthening this connection would yield still further support for the hypothesis that LHF is at least partly responsible for lexical overuse in LLM outputs compared to human-generated text. 

\section{Conclusion}
\label{sec:conclusion}

LHF is known to be a useful tool for aligning the outputs of LLMs more closely with human expectations. Our results, however, suggest that an accidental byproduct of such alignment efforts is lexical overuse. Does the overuse of particular words by LLMs constitute a failure of alignment? And should developers intervene to reduce the prevalence of these words? The answers to both questions depend on whose lexical preferences LLMs ought to reflect. Our research suggests that these models are making lexical choices that align with the preferences and expectations of LHF workers; but these same lexical choices may not satisfy consumers unhappy with LLMs' overuse of words like ``delve.'' 

If intervention is desired, our procedure offers a straightforward way of identifying potential cases of lexical overuse. While some manual verification is still needed, the procedure effectively identifies many of the most extreme instances of potential overuse. Importantly, our findings also highlight one place where interventions could be targeted:\ LHF datasets. Different strategies could be employed. For instance, developers and data scientists could diversify the workforce of human evaluators providing feedback for LHF \cite{sheikh2024delve}, or datasets could be adjusted post-collection to ensure greater balance. 

While we leave open the question of whether intervention is necessary, we note a shift in the dynamics of language change:\ Workers from the Global South are now influencing the language of language technologies, which are subsequently deployed globally. In the past, changes have predominantly flowed in the opposite direction \cite{kwet2019digital,hmensa2024artificial}. However, those who wield this linguistic influence are in positions of economic precarity rather than positions of power.

Finally, our research contributes to the growing body of work on explainable AI \cite{sculley2015hidden,zhao2024explainability,cambria2024xai}:\ Through systematic investigation, meaningful insights into the workings of artificial neural networks can be gained (see also discussion in \cite{templeton2024scaling}). However, a key difficulty for such research is the lack of transparency surrounding LLM development \cite{bommasani2021opportunities}. This includes lack of process transparency, as all major tech companies obscure the details of their LHF procedures, arguably in part to avoid scrutiny of poor working conditions for human evaluators, who are frequently underpaid and stressed \cite{toxtli2021quantifying,roberts2022precarious,novick2023dirty}. Lack of data transparency remains an issue as well, with many LHF datasets not being publicly available. These failures of transparency are worrisome in light of the significant impact that language technology has on global language usage. By facilitating insights like those presented here, publicizing information about model training can aid efforts to align LLMs more closely with human expectations.

%

\appendix

\section{Appendix}
\label{app:appendix}


\noindent \textbf{Permitted Countries}: Bangladesh, Belize, Botswana, Cameroon, Ethiopia, Fiji, Gambia, Ghana, Guyana, Indonesia, Kenya, Liberia, Malawi, Malaysia, Mauritius, Micronesia, Montserrat, Namibia, Nigeria, Pakistan, P.\ N.\ G.,\ Philippines, S.\ Africa, Sri Lanka, Swaziland, Tanzania, Uganda, Zambia, Zimbabwe. 

\vspace{0.2cm}

\noindent \textbf{Words from overuse literature}:\ advancements, aligns, boasts, commendable, comprehending, crucial, delve, delved, delves, delving, emphasizing, garnered, groundbreaking, intricacies, intricate, invaluable, meticulous, meticulously, notable, noteworthy, pivotal, potential, realm, showcases, showcasing, significant, strategically, surpasses, surpassing, underscore, underscores, underscoring. 

\end{document}